\documentclass[letterpaper, 10 pt, conference]{ieeeconf}
\IEEEoverridecommandlockouts                              
\usepackage{graphics} 
\usepackage{amsmath} 
\usepackage{amssymb}  

\usepackage[noend]{algorithmic}
\usepackage[ruled]{algorithm}
\usepackage{mathtools}
\usepackage{pstool}

\usepackage{balance}

\usepackage{todonotes}

\usepackage{pgf}
\usepackage{tikz}
\usetikzlibrary{shapes,arrows,automata,backgrounds,calc}

\DeclareMathOperator*{\argmin}{arg\,min}

\usepackage{subcaption}
\DeclareCaptionLabelSeparator{periodspace}{.\quad}
\definecolor{blue}{rgb}{0,0.4470,0.7410}
\definecolor{red}{rgb}{0.8500,0.3250,0.0980}
\definecolor{yellow}{rgb}{0.9290,0.6940,0.1250}
\definecolor{lila}{rgb}{0.4940,0.1840,0.5560}
\definecolor{green}{rgb}{0.4660,0.6740,0.1880}

\title{\LARGE \bf Model-Based Policy Search for Automatic Tuning of\\Multivariate PID Controllers}

\author{Andreas Doerr$^{1,2}$, Duy Nguyen-Tuong$^{1}$, Alonso Marco$^{2}$, Stefan Schaal$^{2,3}$, Sebastian Trimpe$^{2}$
\thanks{*This research was supported in part by Robert Bosch GmbH, the Max Planck Society, National Science Foundation grants IIS-1205249, IIS-1017134, EECS-0926052, the Office of Naval Research, and the Okawa Foundation.}
\thanks{$^{1}$Bosch Center for Artificial Intelligence, Renningen, Germany.\newline
\{andreas.doerr3, duy.nguyen-tuong\}@de.bosch.com}%
\thanks{$^{2}$Autonomous Motion Department at the Max Planck Institute for Intelligent Systems, T\"ubingen, Germany.\newline
\{amarco, strimpe\}@tue.mpg.de, sschaal@is.mpg.de}%
\thanks{$^{3}$Computational Learning and Motor Control lab at the University of Southern California, Los Angeles, CA, USA.}%
}

\usepackage{fancyhdr}
\newcommand{\mytitle}{\textbf{Accepted final version.}
To appear in \textit{2017 IEEE International Conference on Robotics and Automation}.\\
\copyright 2017 IEEE. Personal use of this material is permitted. Permission from IEEE must be obtained for all other uses, in any current or future media, including reprinting/republishing this material for advertising or promotional purposes, creating new collective works, for resale or redistribution to servers or lists, or reuse of any copyrighted component of this work in other works.}
\begin{document}
\maketitle

\thispagestyle{fancy}	
\pagestyle{empty}

\begin{abstract}
PID control architectures are widely used in industrial applications.
Despite their low number of open parameters, tuning multiple, coupled PID controllers can become tedious in practice.
In this paper, we extend PILCO, a model-based policy search framework, to automatically tune multivariate PID controllers purely based on data observed on an otherwise unknown system.
The system's state is extended appropriately to frame the PID policy as a static state feedback policy.
This renders PID tuning possible as the solution of a finite horizon optimal control problem without further a priori knowledge.
The framework is applied to the task of balancing an inverted pendulum on a seven degree-of-freedom robotic arm, thereby demonstrating its capabilities of fast and data-efficient policy learning, even on complex real world problems.
\end{abstract}

\section{Introduction}
\label{sec:Introduction}

Proportional, Integral and Derivative (PID) control structures are still the main control tool being used in industrial applications, in particular in the process industry \cite{cominos2002pid}, but also in automotive applications \cite{jiang2001application} and in low-level control in robotics \cite{siciliano2010robotics}.
The large share of PID controlled applications is mainly due to the past record of success, the wide availability, and the simplicity in use of this technique.

In practice, control design is still often achieved by tedious manual tuning or by heuristic PID tuning rules \cite{o2009handbook}.
More advanced tuning concepts are most frequently developed for Single-Input-Single-Output (SISO) systems \cite{aastrom2006advanced, berner2016asymmetric}.
For Multi-Input-Multi-Output (MIMO) systems, popular tuning methods, such as biggest log-modulus and the dominant pole placement tuning method \cite{johnson2005}, strive to tune each control loop individually, followed by a collective de-tuning to stabilize the multi-loop system.
These tuning methods, however, rely on linear process models and require stable processes.
For general PID control structures where multiple controllers act on each input, controller design is usually conducted by decoupling the process, subsequently allowing the design of individual SISO PIDs.
One example are online adjusted precompensators, which decouple the process transfer function matrix \cite{yamamoto2004design}.

\begin{figure}
	\centering
	\includegraphics[width=0.65\columnwidth]{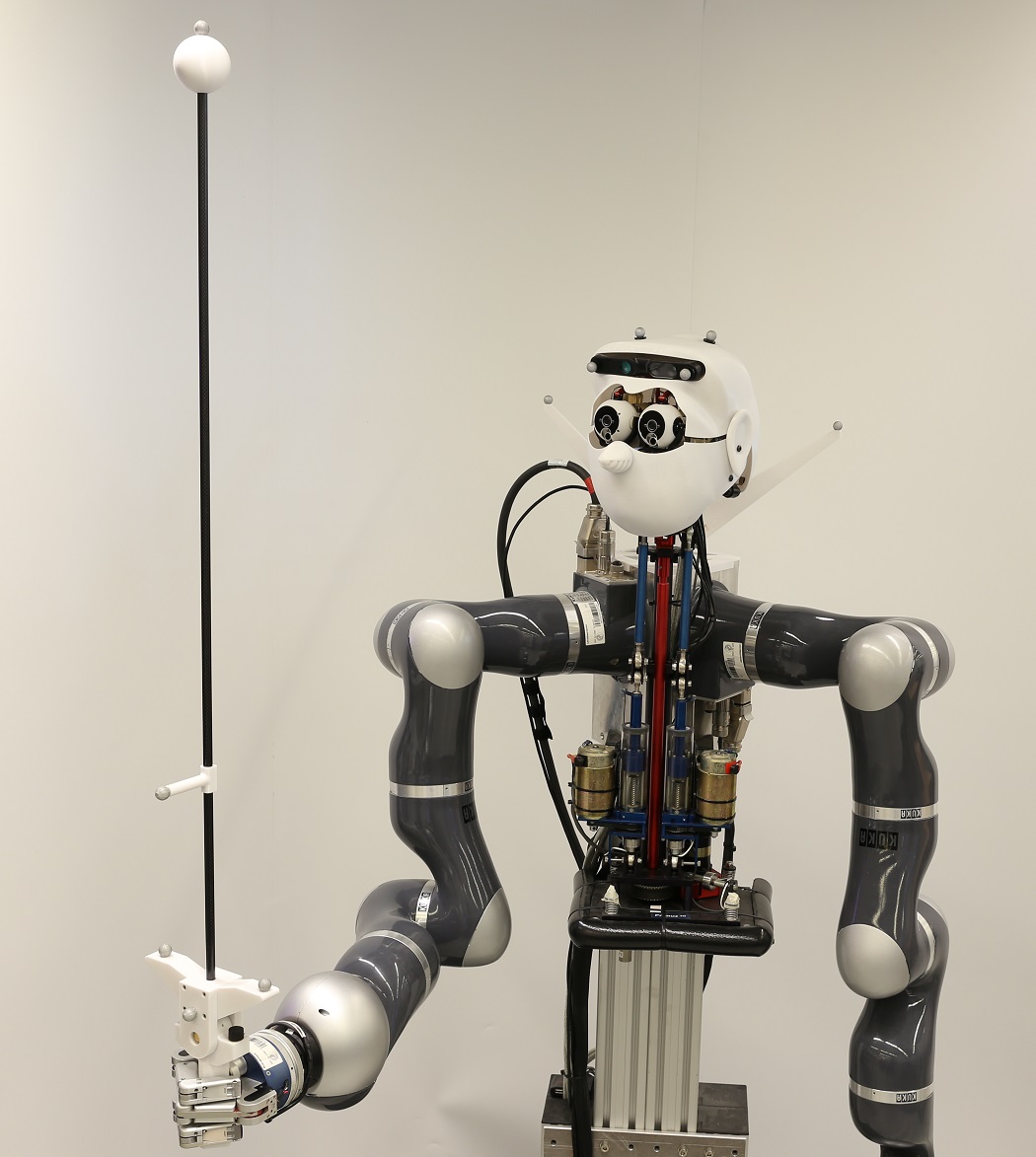}
	\caption{Humanoid upper-body robot, Apollo, balancing an inverted 
	pendulum. Using the proposed framework, coupled PID and PD controllers are trained to stabilize the pole in the central, upright position without requiring prior knowledge of the pendulum system dynamics.}
	\label{fig:ApolloSystem}
\end{figure}

In this paper, we extend Probabilistic Inference for Learning COntrol (PILCO) \cite{deisenroth2011pilco}, a framework for iteratively improving a controller based on its expected finite horizon cost as predicted from a learned system model.
This model is learned in a fully Bayesian setting using Gaussian Process Regression (GPR) as a non-parametric function estimator \cite{kocijan2005dynamic}.
Gaussian Processes (GPs) as probability distributions over a function space do not only provide a prediction but also an uncertainty measure of the process dynamics.
The uncertainty component can be utilized to implement cautious control \cite{murray2003adaptive} or trigger further exploration to improve the model knowledge \cite{murray2002nonlinear}.
As it is based on the full nonlinear system model, this framework takes into account all process couplings for controller tuning.
In contrast to some of the aforementioned PID tuning concepts, e.g. SISO or multi-loop tuning methods, this framework imposes no structural restrictions on the multivariate PID control design.

PILCO has been successfully applied to reinforcement learning tasks, such as inverted pendulum swing-up \cite{deisenroth2011pilco}, control of low-cost manipulators \cite{deisenroth2011learning}, and real-world applications like throttle valve control \cite{bischoff2013learning}. These examples focus mostly on nonlinear state feedback policies. In industrial applications, however, it is desirable to obtain interpretable control designs, which is the case for PID control structures rather than for arbitrarily complex nonlinear control designs.

\textit{Contribution of the paper:} We propose a general framework for multivariate PID controller tuning based on PILCO.
In particular, the developed framework (i) is applicable to nonlinear MIMO systems with arbitrary couplings; (ii) allows for the tuning of arbitrary multivariate PID structures; (iii) involves a probabilistic treatment of recorded system data to iteratively improve control performance without a priori knowledge; and (iv) requires no process model.
The auto-tuning method is demonstrated in pole balancing experiments on Apollo, a complex robot platform as shown in Fig.\,\ref{fig:ApolloSystem}, coping with imperfect low-level tracking controllers and unobserved dynamics.

\textit{Outline of the paper:} The paper continues with an introduction of the policy search problem and PILCO in Sec.\,\ref{sec:ProbabilisticInferenceForControl}.
The proposed multivariate PID tuning framework is then developed in Sec.\,\ref{sec:PolicySearchForPIDControl}.
Section\,\ref{sec:ExperimentalEvaluation} presents the results of applying the framework for tuning coupled PID controllers on Apollo.
The paper concludes with remarks and propositions for future work in Sec.\,\ref{sec:Conclusion}.

\section{Probabilistic Inference for Control}
\label{sec:ProbabilisticInferenceForControl}

In this section, we introduce the policy search problem and notation.
Subsequently, the main ideas of PILCO \cite{deisenroth2011pilco, deisenroth2015gaussian} are briefly discussed.

\subsection{Problem Statement}
\label{sec:ProblemStatement}

We consider discrete time dynamic systems of the form
\begin{align} 
\mathbf{x}_{t+1} &= f(\mathbf{x}_t, \mathbf{u}_t) + \epsilon_t \, ,
\label{eq:GeneralSystemDynamics}
\end{align}
with continuously valued state $\mathbf{x}_t \!\in\! \mathbb{R}^D$ and input $\mathbf{u}_t \!\in\! \mathbb{R}^F$.
The system dynamics $f$ is not known a priori.
We assume a fully measurable state, which is corrupted by zero-mean independent and identically distributed (i.i.d.) Gaussian noise, i.e.\,$\epsilon_t\!\sim\!\mathcal{N}({\bf 0}, \Sigma_\epsilon)$.

One specific reinforcement learning formulation aims at minimizing the expected cost-to-go given by
\begin{equation}
J = \sum_{t=0}^T \mathbb{E}[c(\mathbf{x}_t, \mathbf{u}_t; t)], 
\label{eq:ObjectiveFunction}
\end{equation}
where an immediate, possibly time dependent cost $c(\mathbf{x}_t, \mathbf{u}_t; t)$ penalizes undesired system behavior.
\textit{Policy search} methods optimize the expected cost-to-go $J$ by selecting the best out of a range of policies $\mathbf{u}_t\!=\!\pi(\mathbf{x}_t; \theta)$ parametrized by $\theta$.
Particularly in \textit{model-based} policy search frameworks, a model $\hat{f}$ of the system dynamics $f$ is utilized to predict the system behavior and to optimize the policy.

\subsection{PILCO}
\label{sec:PILCO}

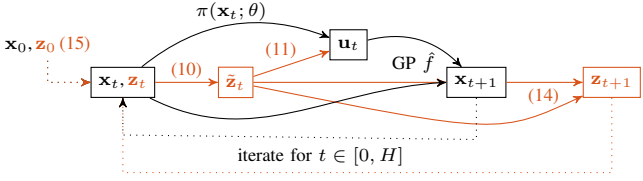
\begin{figure}
\centering
\begin{tikzpicture}[->,>=stealth',font=\fontsize{7}{7}\selectfont]
  \tikzstyle{every state}=[fill=none,rectangle,draw=black,text=black,inner sep=0.1cm,minimum size=0.3cm,minimum height=0.4cm]

  \tikzstyle{label}=[outer sep=0cm,draw=none,inner sep=0.05cm]

  \node[state,draw=none] (X) at(0,1.7)        {$\mathbf{x}_0$,\,\textcolor{red}{$\mathbf{z}_0$\,\eqref{eq:InitialState}}};
..\node[state] (A) at(1,1.2)            {$\mathbf{x}_t$,\,\textcolor{red}{$\mathbf{z}_t$}};
..\node[state, draw=red] (B) at(2.5,1.2)            {\textcolor{red}{$\tilde{\mathbf{z}}_t$}};
..\node[state] (C) at(4,1.7)        {$\mathbf{u}_t$};
..\node[state] (D) at(5.7,1.2)          {$\mathbf{x}_{t+1}$};
..\node[state, draw=red] (E) at(7.5,1.2)            {\textcolor{red}{$\mathbf{z}_{t+1}$}};

  \draw[dotted,->] (X) -- (0,1.2) -- (A);
  \draw[red,dotted,->] (X) -- (0,1.2) -- (A);
  \path[red] (A) edge              node[label, above]{\eqref{eq:StateAugmentation}} (B);
  \path[red] (B) edge [bend left,in=180,out=0]  node[pos=0.35, above]{\eqref{eq:PIDStaticStateFeedback}} (C);
  \path[red] (B) edge node[pos=0.83, above]{\textcolor[rgb]{0,0,0}{GP $\hat{f}$}} (D);
  \draw[red][->] (B) .. controls(6,0.5) .. (E)
            node[pos=0.82,above] {\eqref{eq:NextState}};
  \path (C) edge [bend left] (D);
  \path[red] (D) edge (E);
  \draw[red,dotted,->] (E) -- (7.5,0.0) -- (1,0.0) -- (A);

  \draw[dotted,->] (D) -- (5.7,0.44) -- node[pos=0.44,below]{iterate for $t\in [0,H]$} (1,0.5) -- (A);
  \path (A) edge [bend left] node[label, pos=0.5,above]{$\pi(\mathbf{x}_t; \theta)$}(C);
  \path (A) edge [bend right, in=180] (D);
\end{tikzpicture}
\caption{Computations for one time step in the long-term prediction. Black: original PILCO state propagation steps. Red: Augmented state propagation to accommodate PID policy optimization.}
\label{fig:PropagationSteps}
\end{figure}

PILCO as a specific model-based policy search framework emphasizes data-efficiency and consistent handling of uncertainty when constructing the system dynamics model $\hat{f}$.
To incorporate all available data from policy rollouts (experiments) on the actual system, a Gaussian Process (GP)\,\cite{williams2006gaussian} is utilized as a non-parametric, probabilistic model.

The PILCO framework is outlined in Alg.\,\ref{alg:PILCO}.
In the inner loop, a simulated rollout is conducted based on the dynamics model $\hat{f}$ and the current policy $\pi(\mathbf{x}_t; \theta)$.
The system's state $\mathbf{x}_t$ is propagated over a finite prediction horizon $H$ starting at the system's initial state $\mathbf{x}_0\!\sim\!\mathcal{N}(\mu_0, \Sigma_0)$ as visualized for one time step in Fig.\,\ref{fig:PropagationSteps} in black for the standard PILCO rollout.
The posterior distribution $p(\mathbf{x}_{t+1}|\mathbf{x}_t, \mathbf{u}_t)$ is approximated in each time step by a Gaussian distribution using moment matching \cite{candela2003propagation} yielding an approximately Gaussian marginal distribution of the long-term predictions $p(\mathbf{x}_0), \ldots, p(\mathbf{x}_H)$.
The expected long-term cost \eqref{eq:ObjectiveFunction} of this rollout as well as its gradients with respect to the policy parameters $\theta$ can be computed analytically.
Policy optimization can then be conducted based on this prediction method using standard gradient-based optimization techniques.

Starting with an initial random policy\footnote{A random policy might be a parametrized policy with randomly chosen parameters (cf.\,\cite{deisenroth2011multiple}) or a random input signal exciting the system to generate initial dynamics data.}, the algorithm optimizes the cost-to-go by repeatedly executing the policy on the system, thereby gathering new data and building the dynamics model, subsequently improving the policy iteratively, until the task has been learned.

\begin{algorithm}
	\begin{algorithmic}[1]
		\STATE \textit{Experiment}: Execute random policy;
		\STATE $\quad$ record $\{ \mathbf{x}_t, \mathbf{u}_t\}_{t=1\ldots T}$
		\STATE Train initial GP dynamics model $\mathbf{x}_{t+1} = \hat{f}(\mathbf{x}_t, \mathbf{u}_t)$
		\REPEAT
		\REPEAT
		\STATE \textit{Simulation}: Predict $J(\theta)$ given $\hat{f}$, $\pi(\mathbf{x}_t;\,\theta)$
		\STATE Analytically compute gradient $dJ(\theta)/d\theta$
		\STATE Gradient-based policy update (e.g.\,CG, BFGS)
		\UNTIL{convergence: $\theta^* = \argmin J(\theta)$}
		\STATE \textit{Experiment}: Execute $\pi(\mathbf{x}_t;\,\theta^*)$; record $\{ \mathbf{x}_t, \mathbf{u}_t \}_{t=1 \ldots T}$
		\STATE Update dynamics model $\hat{f}$ using all recorded data
		\UNTIL{task learned}
		\RETURN $\pi^*(\mathbf{x}_t;\,\theta^*)$
	\end{algorithmic}
	\caption{PILCO}
	\label{alg:PILCO}
\end{algorithm}

\section{Policy Search for PID Control}
\label{sec:PolicySearchForPIDControl}

We first introduce the general description of the considered PID control structure in Sec.\,\ref{sec:PIDControlNetworks}; the necessary modifications to the policy search framework are derived in Sec.\,\ref{sec:SystemStateAugmentation} to Sec.\,\ref{sec:StatePropagation}, followed by an analytic derivation of the required cost function gradients in Sec.\,\ref{sec:CostFunctionDerivatives}.

\subsection{PID Control Policies}
\label{sec:PIDControlNetworks}

The control output of a scalar PID controller is given by
\begin{align}
u_t &= K_\text{p} e_t + K_\text{i} \int_{0}^{t \cdot \Delta T} e(\tau) d\tau + K_\text{d} \dot{e}_t
\label{eq:PIDController} \\
e_t &= x_{\text{des},t} - x_t\,,
\label{eq:Error}
\end{align}
where $e(\tau)$ denotes the continuous error signal.
The current desired state $x_{\text{des},t}$ can be either a constant set-point or a time-variable goal trajectory.
This controller is agnostic to the system dynamics and depends only on the system's error.
Each controller is parametrized by its proportional, integral and derivative gain ($\theta_\text{PID}\!=\!(K_\text{p}, K_\text{i}, K_\text{d})$).

A general PID control structure $C(s)$ for MIMO processes \eqref{eq:GeneralSystemDynamics} can be described in transfer function notation by a $D \times F$ transfer function matrix
\begin{equation}
C(s)
=
\begin{bmatrix}
c_{11}(s) & \cdots & c_{1D}(s) \\
\vdots    & \ddots & \vdots \\
c_{F1}(s) & \cdots & c_{FD}(s) \\
\end{bmatrix}\,,
\label{eq:GeneralController}
\end{equation}
where $s$ denotes the complex Laplace variable and $c_{ij}(s)$ are of PID type.
The multivariate error is given by $\mathbf{e}_t\!=\!\mathbf{x}_{\text{des},t}\!-\!\mathbf{x_t}\!\in\!\mathbb{R}^D$ such that the multivariate input becomes $\mathbf{u}(s)\!=\!C(s)\mathbf{e}(s)$.
PID literature typically distinguishes between tuning methods for \textit{multi-loop} PID control and \textit{multivariable} PID control systems. The former have a diagonal transfer function matrix $C(s)$ whereas the latter allows PID controllers on all elements of $C(s)$, i.e.\,all combinations of errors and inputs can be controlled, thus allowing additional cross couplings.

\begin{figure}
\centering
\begin{tikzpicture}[->,>=stealth',font=\fontsize{7}{7}\selectfont]
  \tikzstyle{every state}=[fill=none,rectangle,draw=black,text=black,inner sep=0.1cm,minimum size=0.3cm,minimum height=0.4cm]

  \tikzstyle{system} = [minimum width=1.5cm, text centered, minimum height=1.3cm, draw=black, rectangle, inner sep=0cm]
  \tikzstyle{policy} = [minimum width=0.8cm, text centered, minimum height=0.5cm, draw=black, rectangle, inner sep=0cm]
  \tikzstyle{input} = [draw=none, rectangle, inner sep=0.1cm]

  \tikzstyle{sum} = [draw=black, circle, minimum width=0.2cm, inner sep=0cm]

  \node (sys1) at(2.7,0.4) [system] {Sys 1};
  \node (pid1) at(1,0.8) [policy] {$\text{PID}_1$};
  \node (pid2) at(1,0) [policy] {$\text{PID}_2$};
  \node (e1) at(0,0.8) [input] {$e_1$};
  \node (e2) at(0,0) [input] {$e_2$};

  \draw[->] (e1) -- (pid1);
  \draw[->] (pid1) -- node[pos=0.6, above]{$u_1$} ++(0.95,0);
  \draw[->] (e2) -- (pid2);
  \draw[->] (pid2) -- node[pos=0.6, above]{$u_2$} ++(0.95,0);

  \node (sys2) at(7.5,0.4) [system] {Sys 2};
  \node (sum) at(6,0.4) [sum] {+};
  \node (pid3) at(5,0.8) [policy] {$\text{PID}_1$};
  \node (pid4) at(5,0) [policy] {$\text{PID}_2$};
  \node (e3) at(4,0.8) [input] {$e_1$};
  \node (e4) at(4,0) [input] {$e_2$};

  \draw[->] (e3) -- (pid3);
  \draw[->] (pid3) -- (6,0.8) -- (sum);
  \draw[->] (e4) -- (pid4);
  \draw[->] (pid4) -- (6,0) --(sum);
  \draw[->] (sum) -- node[pos=0.6, above]{$u$} (sys2);

\end{tikzpicture}
\caption{Possible PID structures, exemplified with two controllers. Left: Individual PID controllers acting on different system inputs. Right: Combination of PID controllers acting on the same system input.}
\label{fig:PIDStructures}
\end{figure}
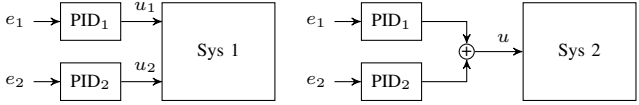

With this framework, we address the general class of \textit{multivariable} PID control systems with no restrictions on the elements of $C(s)$.
Examples for possible PID structures are shown in Fig.\,\ref{fig:PIDStructures}.

\subsection{System State Augmentation}
\label{sec:SystemStateAugmentation}

We present a sequence of state augmentations such that any multivariable PID controller \eqref{eq:GeneralController} can be represented as a parametrized static state feedback law.
A visualization of the state augmentation integrated into the one-step-ahead prediction is shown in red in Fig.\,\ref{fig:PropagationSteps} in comparison with the standard PILCO setting (in black).
Given a Gaussian distributed initial state $\mathbf{x}_0$, the resulting predicted states will remain Gaussian for the presented augmentations.

To obtain the required error states for each controller \eqref{eq:PIDController}, we define a new system state $\mathbf{z}_t$ that keeps track of the error at the previous time step and the accumulated error,
\begin{equation}
\mathbf{z}_t := (\mathbf{x}_t, \mathbf{e}_{t-1}, \Delta T \sum_{\tau=0}^{t-1}  \mathbf{e}_\tau)\,,
\label{eq:NewSystemState}
\end{equation}
where $\Delta T$ is the system's sampling time.
For simplicity, we denote vectors as tuples $(\mathbf{v}_1, \ldots, \mathbf{v}_n)$, where $\mathbf{v}_i$ may be vectors themselves.
The following augmentations are made to obtain the necessary policy inputs:

\subsubsection{Desired Goal State}
\label{sec:DesiredGoalState}

The desired set-point or target trajectory state\footnote{Drawing the desired state from a Gaussian distribution yields better generalization to unseen targets as discussed in \cite{deisenroth2011learning}.} $\mathbf{x}_{\text{des},t}\!\sim\!\mathcal{N}(\mu_{\text{des},t}, \Sigma_{\text{des},t})$ is independent of $\mathbf{z}_t$ yielding:
\begin{equation}
\begin{bmatrix}
\mathbf{z}_t \\
\mathbf{x}_\text{des,t}
\end{bmatrix}
\sim \mathcal{N}\left(
\begin{bmatrix}
\mu_z \\
\mu_{\text{des},t}
\end{bmatrix},
\begin{bmatrix}
\Sigma_z & \mathbf{0} \\
\mathbf{0} & \Sigma_{\text{des},t}
\end{bmatrix}\right)\,.
\label{eq:DesiredStateAugmentation}
\end{equation}

\subsubsection{Error States}
\label{sec:ErrorStates}

The current error is a linear function of $\mathbf{z}_t$ and $\mathbf{x}_{\text{des},t}$.
The current error derivative and integrated error are approximated by
\begin{align}
\dot{\mathbf{e}}_t &\approx \frac{\mathbf{e}_t - \mathbf{e}_{t-1}}{\Delta T}\,,
\label{eq:DerivativeError}\\
\int_{0}^{t \cdot \Delta T} \mathbf{e}(\tau) d\tau &\approx \Delta T \sum_{\tau=0}^{t-1} \mathbf{e}_\tau + \Delta T \mathbf{e}_t\,.
\label{eq:IntegralError}
\end{align}
Both approximations are linear transformations of the augmented state.
The resulting augmented state distribution remains Gaussian as it is a linear transformation of a Gaussian random variable (see Appendix).

The computation of the derivative error is prone to measurement noise.
Yet, this framework can readily be extended to incorporate a low-pass filtered error derivative, which we omit for notational simplicity.
In this case, additional historic error states would be added to the state $\mathbf{z}_t$ to provide the input for a low-pass Finite Impulse Response (FIR) filter.

The fully augmented state is given by
\begin{equation}
\tilde{\mathbf{z}}_t := (\mathbf{z}_t,\,\mathbf{x}_{\text{des},t},\,\mathbf{e}_t,\,\Delta T \sum_{\tau=0}^{t} \mathbf{e}_\tau, \frac{\mathbf{e}_t - \mathbf{e}_{t-1}}{\Delta T})\,.
\label{eq:StateAugmentation}
\end{equation}

\subsection{PID as Static State Feedback}
\label{sec:PIDAsStaticStateFeedback}

Based on the augmented state $\tilde{\mathbf{z}}_t$, the PID control policy for multivariate controllers can be expressed as a static state feedback policy:
\begin{align}
\begin{split}
\mathbf{u}_t &= \mathbf{A}_\text{PID} (\tilde{\mathbf{z}}_t^{(3)}, \tilde{\mathbf{z}}_t^{(4)}, \tilde{\mathbf{z}}_t^{(5)})\\
&= \mathbf{A}_\text{PID} \left(\mathbf{e}_t, 
\Delta T \sum_{\tau=0}^t \mathbf{e}_t, \frac{\mathbf{e}_t - \mathbf{e}_{t-1}}{\Delta T}\right)\,,
\end{split}
\label{eq:PIDStaticStateFeedback}
\end{align}
where $\tilde{\mathbf{z}}_t^{(i)}$ indicates the i-th term of \eqref{eq:StateAugmentation}.
The specific structure of the multivariate PID control law is defined by the parameters in $\mathbf{A}_\text{PID}$.
For example, PID structures as shown in Fig.\,\ref{fig:PIDStructures} would be represented by
\begin{align}
\mathbf{A}_\text{left} &= 
\begin{bmatrix}
K_{\text{p,1}}& 0 & K_{\text{i,1}}& 0 & K_{\text{d,1}} & 0 \\
0 & K_{\text{p,2}} & 0 & K_{\text{i,2}} & 0 & K_{\text{d,2}}
\end{bmatrix}\,,\\
\mathbf{A}_\text{right} &= 
\begin{bmatrix}
K_{\text{p,1}} & K_{\text{p,2}} & K_{\text{i,1}} & K_{\text{i,2}} & K_{\text{d,1}} & K_{\text{d,2}}
\end{bmatrix}\,.
\label{eq:PIDParametrization}
\end{align}

\subsection{State Propagation}
\label{sec:StatePropagation}

Given the Gaussian distributed state and control input as derived in Sec.\,\ref{sec:SystemStateAugmentation} and Sec.\,\ref{sec:PIDAsStaticStateFeedback}, the next system state is computed using the GP dynamics model $\hat{f}$.
PILCO approximates the predictive distribution $p(\mathbf{x}_{t+1})$ by a Gaussian distribution using exact moment matching.
From the dynamics model output $\mathbf{x}_{t+1}$ and the current error stored in $\tilde{\mathbf{z}}_t$, the next state is obtained as
\begin{equation}
\mathbf{z}_{t+1} = (\mathbf{x}_{t+1}, \tilde{\mathbf{z}}_t^{(3)}, \tilde{\mathbf{z}}_t^{(4)}) = (\mathbf{x}_{t+1},\,\mathbf{e}_t,\,\Delta T \sum_{\tau=0}^{t} \mathbf{e}_\tau)\,.
\label{eq:NextState}
\end{equation}
Iterating \eqref{eq:NewSystemState} to \eqref{eq:NextState}, the long-term prediction can be computed over the prediction horizon $H$ as shown in Fig.\,\ref{fig:PropagationSteps}.
For the initial state, we define
\begin{equation}
\mathbf{z}_0 := (\mathbf{x}_0,\,\mathbf{x}_{\text{des},0}-\mathbf{x}_0,\,\mathbf{0})\,.
\label{eq:InitialState}
\end{equation}

\subsection{Cost Function Derivatives}
\label{sec:CostFunctionDerivatives}

Given the presented augmentation and propagation steps, the expected cost gradient can be computed analytically such that the policy can be efficiently optimized using gradient-based methods.
We summarize the high-level policy gradient derivation steps to point out the modifications to standard PILCO that are necessary to allow PID policy optimization.
The expected cost\footnote{We only address cost on the state $\mathbf{z}_t$ for simplicity. For practical implementations, cost on $\mathbf{u}$ can be included by adding past inputs into the system state as shown in Sec.\,\ref{sec:ExperimentalEvaluation}.} derivative is obtained as
\begin{equation}
\frac{dJ(\theta)}{d\theta} = \sum_{t=1}^H \frac{d}{d\theta} 
\underbrace{\mathbb{E}_{\mathbf{z}_t}[c(\mathbf{z}_t)]}_{=:\mathcal{E}_t} 
=
\sum_{t=1}^T
\frac{d\mathcal{E}_t}{d p(\mathbf{z}_t)}
\frac{d p(\mathbf{z}_t)}{d\theta}\,.
\label{eq:CostFunctionGradient}
\end{equation}
To simplify the notation, we write $d p(\mathbf{z}_t)$ to denote the sufficient statistics derivatives $d\mu_t$ and $d\Sigma_t$ of a Gaussian random variable $p(\mathbf{z}_t)\!\sim\!\mathcal{N}(\mu_t, \Sigma_t)$ (analogous to the treatment in \cite{deisenroth2015gaussian}).
The gradient of the immediate loss with respect to the state distribution, $d\mathcal{E}_t/d p(\mathbf{z}_t)$, is readily available for most standard cost functions like quadratic or saturated exponential terms and Gaussian input distributions (cf.\,\cite{deisenroth2011learning}).
The gradient for each predicted state in the long-term rollout is obtained by applying the chain rule to \eqref{eq:NextState} resulting in
\begin{align}
\frac{d p(\mathbf{z}_{t+1})}{d\theta} = 
\textcolor{blue}{\frac{\delta p(\mathbf{z}_{t+1})}{\delta p(\tilde{\mathbf{z}}_{t})}}
\frac{d p(\tilde{\mathbf{z}}_{t})}{d\theta} 
+
\textcolor{blue}{\frac{\delta p(\mathbf{z}_{t+1})}{\delta p(\mathbf{x}_{t+1})}}
\frac{d p(\mathbf{x}_{t+1})}{d\theta}\,.
\label{eq:StateConcatenationDerivativeRule}
\end{align}
The derivatives highlighted in blue are computed for the linear transformation in \eqref{eq:NextState} according to the general rules for linear transformations on Gaussian random variables as summarized in the appendix.
Based on the dynamics model prediction as detailed in Sec.\,\ref{sec:StatePropagation}, the gradient of the dynamics model output $\mathbf{x}_{t+1}$ is given by
\begin{align}
\frac{d p(\mathbf{x}_{t+1})}{d\theta} = 
\textcolor{red}{\frac{\delta p(\mathbf{x}_{t+1})}{\delta p(\tilde{\mathbf{z}}_{t})}}
\frac{d p(\tilde{\mathbf{z}}_{t})}{d\theta} 
+
\textcolor{red}{\frac{\delta p(\mathbf{x}_{t+1})}{\delta p(\mathbf{u}_{t})}}
\frac{d p(\mathbf{u}_{t})}{d\theta}\,.
\label{eq:PredictionDerivativeRule}
\end{align}
The derivatives shown in red can be computed analytically for the specific dynamics model \cite{candela2003propagation}.
Applying the chain rule for the policy output $p(\mathbf{u}_{t})$ obtained by \eqref{eq:PIDStaticStateFeedback} yields
\begin{equation}
\frac{d p(\mathbf{u}_{t})}{d\theta}
=
\textcolor{blue}{\frac{\delta p(\mathbf{u}_{t})}{\delta p(\tilde{\mathbf{z}}_{t})}}
\frac{d p(\tilde{\mathbf{z}}_{t})}{d \theta}
+
\textcolor{blue}{\frac{\delta p(\mathbf{u}_{t})}{\delta \theta}}\,,
\label{eq:PolicyDerivativeRule}
\end{equation}
The derivatives marked in blue are introduced by the linear control law \eqref{eq:PIDStaticStateFeedback} and can be computed as summarized in the appendix.
The gradient of the augmented state is given by
\begin{align}
\frac{d p(\tilde{\mathbf{z}}_{t})}{d\theta}
=
\textcolor{blue}{\frac{d p(\tilde{\mathbf{z}}_{t})}{d p(\mathbf{z}_{t})}}
\frac{d p(\mathbf{z}_{t})}{d\theta}\, 
\label{eq:StateAugmentationDerivativeRule}
\end{align}
Again, the part marked in blue is computed for the linear transformation \eqref{eq:StateAugmentation}.
Starting from the initial state where $dp(\mathbf{z}_0)/d\theta\!=\!0$, we obtain the gradients for all states with respect to the policy parameters $d p(\mathbf{z}_{t})/ d\theta$ by iteratively applying \eqref{eq:StateConcatenationDerivativeRule} to \eqref{eq:StateAugmentationDerivativeRule} for all time steps $t$.

\section{Experimental Evaluation}
\label{sec:ExperimentalEvaluation}

To demonstrate the capabilities of the presented framework to automatically tune coupled PID controllers without prior system knowledge and in a data-efficient fashion, we consider the problem of balancing an inverted pendulum on the Apollo robot as shown in Fig.\,\ref{fig:ApolloSystem}.
The inverted pendulum is a well-known benchmark in the control and reinforcement learning communities \cite{sutton1998reinforcement,anderson1989learning}.
Demonstrations of the iterative learning process and the resulting optimized policy can be found in the supplementary video material.

\subsection{Experimental Setup}
\label{sec:ExperimentalSetup}

We employ an imperfect inverse dynamics model of Apollo's seven Degree-of-Freedom (DoF) robotic arm to compute the joint torques necessary to track the desired end effector acceleration $u_t$ \cite{righetti2014autonomous}.
Technical details concerning the hardware platform can be found in \cite{marco_ICRA_2016}, where reinforcement learning on this platform has been addressed using Bayesian Optimization techniques.
The state of the system $\mathbf{x}_t$ comprises the end effector position $x$, velocity $\dot{x}$, pendulum angle $\phi$, and angular velocity $\dot{\phi}$.

\begin{figure}
	\centering
	\includegraphics{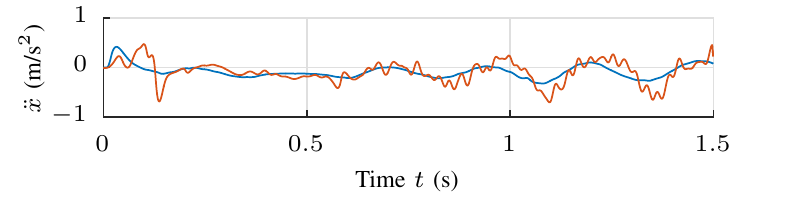}
	\caption{Comparison of commanded (blue) and executed (red) end effector acceleration for a policy rollout on Apollo. Additionally to the tracking errors introduced by the joint level controllers, the system shows further deterioration in acceleration tracking due to errors in the inverse dynamics model and due to noise and delays.}
	\label{fig:AccelerationInputComparison}
\end{figure}

During the rollouts, the commanded acceleration is limited to $u_{\text{max}}\!=\!3$\,m/s$^2$ for safety reasons.
Test policies are executed for 20\,s or interrupted once safety limits ($x_{\text{max}}\!=\!0.3$\,m, $\theta_{\text{max}}\!=\!30^\circ$) are violated.
The control signal is computed at 100\,Hz and low-pass filtered by a second order Butterworth filter having a cut-off frequency of 20\,Hz.

The policy is optimized on a prediction horizon of $T\,=\,10\,s$ based on a saturated loss function \cite{deisenroth2011pilco} given by
\begin{equation}
c(\mathbf{e}_t, \bar{\mathbf{u}}_t) = 1 - \exp(-(\mathbf{e}_t^T Q \mathbf{e}_t + \bar{\mathbf{u}}_t^T R \bar{\mathbf{u}}_t)/2)\,.
\label{eq:SaturatedLossFunction}
\end{equation}
For balancing the pendulum in the upright position the desired trajectory is given by $\mathbf{x}_{\text{des},t} = \mathbf{0}$.
Weights are set to $ Q = \text{diag}(1/0.2^2, 1/0.02^2)$ for end effector and pendulum position error, and to $R = 1/0.4^2$ for the control input.
The selected cost function saturates quickly for $x\!>\!w$ if $x$ is weighted by $1/w^2$.
End effector position and input are therefore only penalized lightly,
which permits higher gains while the pendulum angle is stabilized as it is penalized much stronger.

\subsection{PID Control Structure Setup}
\label{sec:PIDControlStructureSetup}

We employ a PID controller on the position error $e_\text{x} = x_{\text{des},t} - x$ and a PD controller acting on the pendulum angular error $e_\theta = \theta_{\text{des},t} - \theta$.
The resulting control structure is shown in the right plot of Fig.\,\ref{fig:PIDStructures}.
The PID/PD control structure with integral control on the end effector position serves to correct for any static bias in the angle measurement (e.g., from imperfect calibration) as is explained in \cite[p.~67]{trimpe2012balancing}.
This structure has successfully been used for other balancing problems \cite{marco_ICRA_2016, trimpe2012balancing}.

The specific structure is chosen based on prior knowledge about the problem at hand as detailed in \cite{marco_ICRA_2016}.
The integrator's contribution is required to counteract any pendulum angle measurement bias introduced by imperfect sensor calibration.
The policy parametrization is therefore given by $\theta\!=\!(K_{\text{p,x}}, K_{\text{i,x}}, K_{\text{d,x}}, K_{\text{p},\theta}, K_{\text{d},\theta})$.
Assuming no prior knowledge, we initialize the policy to zero.
Both controllers are coupled by the system dynamics and can therefore not be tuned independently, which makes the inverted pendulum a well suited benchmark for multivariate PID controller tuning.

\subsection{Modifications for Dynamics Model Learning}
\label{sec:SetupForDynamicsModelLearning}

When first training GP dynamics models $\mathbf{x}_{t+1} = \hat{f}(\mathbf{x}_t, \mathbf{u}_t)$ in the standard way (Sec.\,\ref{sec:ProbabilisticInferenceForControl}), this did not lead to acceptable models for long-term predictions and thus successful controller learning.
The problems are caused by imperfections in the inverse dynamics model, joint friction and stiction, as well as sensor and actuator delay.
These factors add unobserved states and therefore additional dynamics to the system, corrupting the measured data.
This is visible in Fig.\,\ref{fig:AccelerationInputComparison}, where the desired acceleration and the numerically computed actual acceleration are visualized for a policy rollout on the robot.
Several adaptations to the standard GP dynamics model learning framework were required to obtain a good prediction model, which we explain next.

\subsubsection{Gaussian Process Setup}
\label{sec:GaussianProcessSetup}

In contrast to the model in \eqref{eq:GeneralSystemDynamics}, we train the GP models to predict the difference between the current and the next state $\Delta \mathbf{x}_t = \mathbf{x}_{t+1} - \mathbf{x}_t$.
We compute a sparse GP using Snelson's approximation \cite{snelson2005sparse} having a covariance parametrized by 400 inducing pseudo-input points.
For both GP models, hyperparameters $\theta_{\text{GP},i} = (l_1, l_2, \sigma_\text{f}, \sigma_\text{n})$ including lengthscales for each dimension, as well as signal and noise variance have to be chosen.
We chose the maximum likelihood estimate (MLE) on the basis of the previously gathered system data to compute the hyperparameters.
These hyperparameters are kept constant during the iterative policy learning.

\subsubsection{NARX Dynamics Model}
\label{sec:NARXDynamicsModel}

Instead of modeling the system's full, four dimensional state and its dynamics, two independent GPs are trained to model the dynamics of the measured state parts; the end effector and pendulum position.
The missing information about the system's velocities and potential latent states is recovered by employing a Nonlinear AutoRegressive eXogenous model (NARX) \cite{billings2013nonlinear} of the form
\begin{equation}
\mathbf{x}_{t+1}\!=\!\hat{f}(\mathbf{x}_t,\ldots,\mathbf{x}_{t-n},\mathbf{u}_t,\ldots,\mathbf{u}_{t-m})\,.
\end{equation}
Information on latent states is implicitly encoded in the measured historic states and inputs.
Different lengths of history might be required for individual parts of the system's state depending on the dynamics' time scales.
We optimize the number of historic states individually for end effector position, pendulum position and control input.
The NSGA II optimizer \cite{deb2002fast} is employed to compute the pareto front of the model's prediction error (using the same dataset as previously used for hyperparameter tuning) and the size of the NARX history.
For our problem, we ended up with a new state of dimensionality 14 given by $\mathbf{x}_\text{NARX} := (x_t,\ldots,x_{t-3},\phi_t,\ldots,\phi_{t-2},u_t,\ldots,u_{t-6})$.
The hidden dynamics between commanded input and executed acceleration (cf.\,Fig.\,\ref{fig:AccelerationInputComparison}) requires a longer history in the input state to capture all relevant effects.

\subsubsection{Data Preprocessing}
\label{sec:DataPreprocessing}

The recorded data is downsampled to 25\,Hz and non-causally low-pass filtered with a 2nd order Butterworth filter and cut-off frequency 12.5\,Hz.
The downsampled control input is obtained by averaging the input signal on each sampling interval.

\subsection{PID Learning Results on Apollo}
\label{sec:PIDLearningResults}

\begin{figure}
	\centering
	\begin{subfigure}{0.32\columnwidth}
		\includegraphics{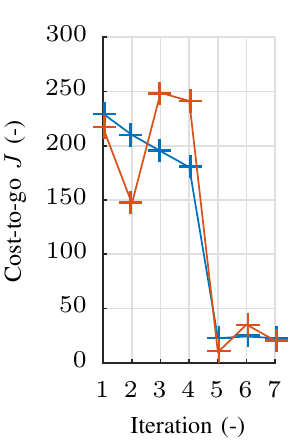}
	\end{subfigure}
	\begin{subfigure}{0.62\columnwidth}
		\includegraphics{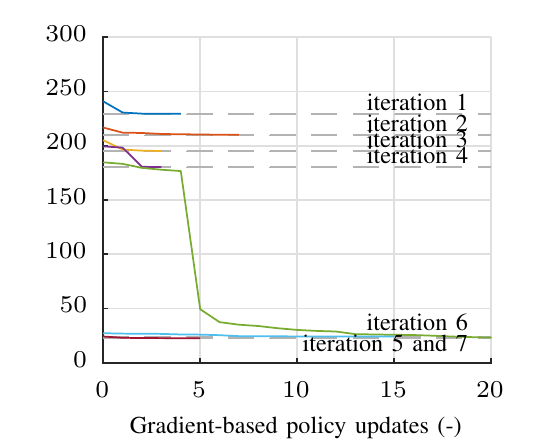}
	\end{subfigure}
	\caption{Expected cost-to-go optimization results. Left: Iterative improvement in predicted loss (blue) compared to the cost observed in a single robot experiment (red). Right: Optimization results for each iteration.}
	\label{fig:LossOptimization}
\end{figure}
\begin{figure*}
	\centering
	\begin{subfigure}[t]{0.32\textwidth}
		\includegraphics{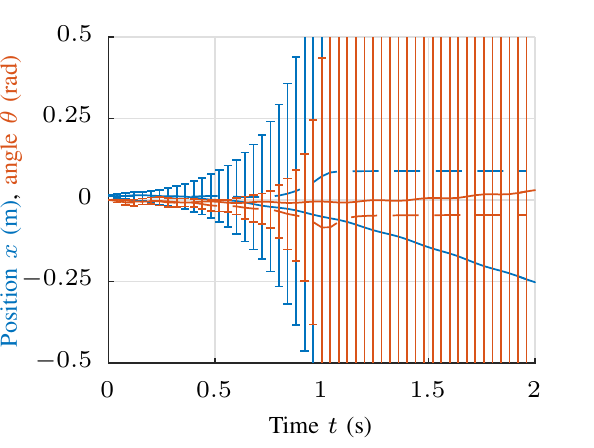}
		\caption{Iteration 1}
		\label{fig:first_iteration_prediction_robot_comparison}
	\end{subfigure}
	\begin{subfigure}[t]{0.32\textwidth}
		\includegraphics{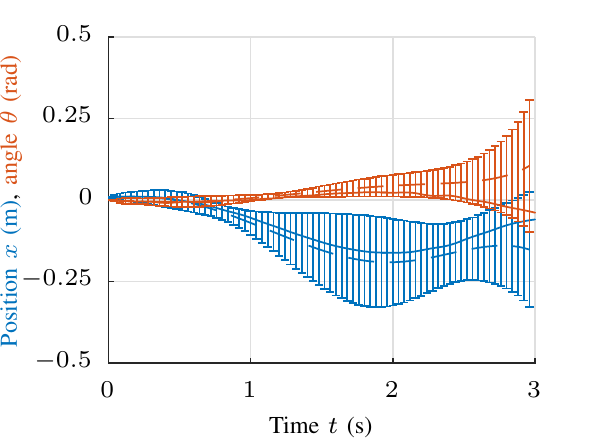}
		\caption{Iteration 3}
		\label{fig:intermediate_iteration_prediction_robot_comparison}
	\end{subfigure}
	\begin{subfigure}[t]{0.32\textwidth}
		\includegraphics{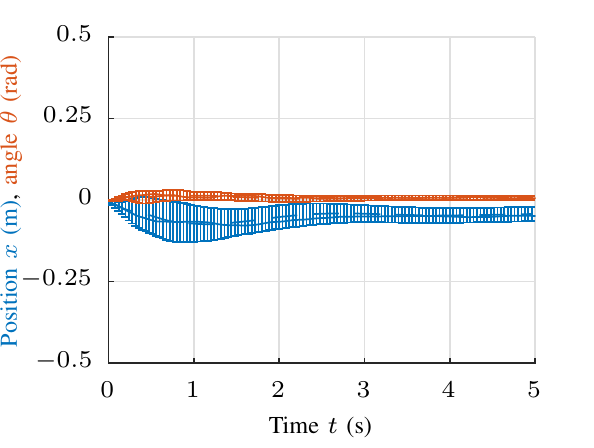}
		\caption{Iteration 7}
		\label{fig:final_iteration_prediction_robot_comparison}
	\end{subfigure}
	\caption{Predicted system behavior (dashed lines, error-bars indicate 95\,\% confidence intervals) and experimental results (solid lines) visualized for the first (a), an intermediate (b) and the final iteration (c) of the policy learning process. The end effector position (blue) and the angular position (red) are shown.}
	\label{fig:PredictionRolloutComparison}
\end{figure*}
\begin{figure}
	\centering
	\includegraphics{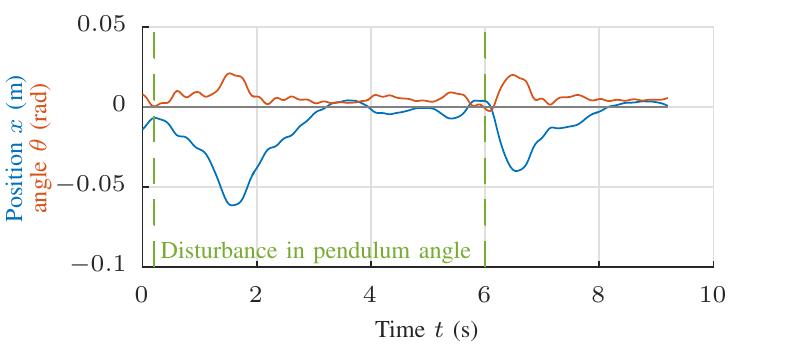}
	\caption{Disturbance rejection of the optimized PID policy. End effector position (blue) and pendulum angle (red) display the closed loop response of the optimized PID policy to manually introduced disturbances.}
	\label{fig:DisturbanceRejection}
\end{figure}
The iterative learning experiment is visualized in Fig.\,\ref{fig:LossOptimization}.
To gather initial data about the system, four random rollouts are conducted, applying a white noise input $u_{\text{rnd},t}\!\sim\!\mathcal{N}(0, 1$\,m/s$^2)$ to the system.
The first iteration is based on the dynamics model learned from these random rollouts.
The left figure visualizes the optimized predicted loss for each iteration in comparison to the actual loss obtained from evaluating \eqref{eq:SaturatedLossFunction} on one sample rollout of the actual system.
The drop of the predicted loss in iteration 5 shows that by that time, sufficient data about the system dynamics has been gathered to find a stabilizing policy parametrization.
In the right figure, the predicted loss optimization is visualized for each iteration individually, as a function of the number of linesearches conducted by the BFGS optimizer.

In Fig.\,\ref{fig:PredictionRolloutComparison}, we visualize the dynamics learning progress, showing predicted system behavior and actual rollout for the dynamics models and policies as obtained at the first, intermediate and final stage of the iterative learning process.
The iterative improvement in model prediction accuracy and the improvement in the PID policy is clearly visible.
The final dynamics model is accurately predicting the stabilization of the system by the optimized policy.

In this example, the total interaction time with the physical system is only 106\,seconds, demonstrating fast and data-efficient learning.
This model-based method outperforms a model-free, Bayesian optimization method (cf.\,\cite{marco_ICRA_2016}), with respect to the number of rollouts on the actual system.
The policy optimization itself is carried out offline, and the predicted system behavior can be utilized to set appropriate safety boundaries to test new controllers without damaging the system.

To demonstrate the robustness of the learned policy, the system is manually deflected.
Disturbances in pendulum angle (cf.\,Fig.\,\ref{fig:DisturbanceRejection}) and end effector position are dispelled fast and without overshoot.
By commanding a non-zero desired trajectory, the learned PID controller can be utilized for tracking tasks as demonstrated in the supplementary video for a sinusoidal end-effector trajectory.

\balance

\section{Conclusion}
\label{sec:Conclusion}

The proposed framework for multivariate PID tuning is flexible with respect to possible PID control structures and process dynamics.
In particular, it is able to cope with general nonlinear MIMO processes and multivariate PID structures.
The presented framework can readily be extended to cascaded PID structures and tracking controllers by considering multiple different \cite{deisenroth2011multiple} and time-varying goal states.

Appropriately dealing with hidden and low-level dynamics, problems found in almost all real-world applications, were major hurdles in the experimental application.
In particular, we found that learning of dynamics models geared towards long-term predictions is key to successful finite horizon policy optimization, but largely unaddressed by current approaches.
The presented dynamics model structure and learning framework helps to alleviate this problem.
Principled ways for improving the learning of long-term prediction models shall be addressed in future work.

\addtolength{\textheight}{-7cm}   

\section*{APPENDIX}
\label{sec:Appendix}

A linear transformation of a Gaussian random variable $\mathbf{X} \sim \mathcal{N}(\mu_X, \Sigma_X) \in \mathbb{R}^D$ is given by
\begin{equation}
\mathbf{Y} = \mathbf{A} \mathbf{X} + \mathbf{b} = \mathcal{N}(\mu_Y, \Sigma_Y) \in \mathbb{R}^P,
\label{eq:LinearTransformation}
\end{equation}
where $\mathbf{A} \in \mathbb{R}^{P \times D}$ and $\mathbf{b} \in \mathbb{R}^{P \times 1}$.
The joint probability distribution of $\mathbf{X}$ and $\mathbf{Y}$ is given by
\begin{equation}
\begin{bmatrix}
\mathbf{X} \\
\mathbf{Y}
\end{bmatrix}
\sim
\mathcal{N}(
\begin{bmatrix}
\mathbf{\mu}_X \\
\mathbf{\mu}_Y
\end{bmatrix}
,
\begin{bmatrix}
\mathbf{\Sigma}_X & \mathbf{\Sigma}_X\mathbf{C}\\
\mathbf{C}^T\mathbf{\Sigma}_X^T & \mathbf{\Sigma}_Y
\end{bmatrix})
\end{equation}
where
\begin{equation}
\mu_Y = \mathbf{A} \mu_X + \mathbf{b},\quad \Sigma_Y = \mathbf{A} \Sigma_X 
\mathbf{A}^T,\quad \mathbf{C} = \mathbf{A}^T
\end{equation}

The non-zero partial derivatives of $Y$'s sufficient statistics are given by
\begin{align}
\frac{\delta \mu_Y}{\delta \mu_X} &= A \in \mathbb{R}^{P\times D}, & \frac{
\delta \Sigma_Y}{\delta \Sigma_X} &= A \otimes A  \in \mathbb{R}^{P^2 \times D^2}\\
\frac{\delta \mu_Y}{\delta A} &= \mu_X^T \otimes I \in \mathbb{R}^{P\times DP}
, & \frac{\delta \mu_Y}{\delta b} &= I \in \mathbb{R}^{P\times P}
\end{align}
\begin{align}
\frac{\delta C_{kl}}{\delta A_{ij}} &= \delta_{il} \delta_{kj},\frac{
\delta C}{\delta A} \in \mathbb{R}^{PD\times D^2},\frac{\delta (\Sigma_{Y})}{\delta A} 
\in \mathbb{R}^{P^2\times PD}\\
\frac{\delta (\Sigma_{Y})_{kl}}{\delta A_{ij}} &= \delta_{lj}(A^T\Sigma_X)_{ki
} + \delta_{kj}(\Sigma_X A)_{il}
\end{align}
where $\otimes$ is the Kronecker product and $\delta_{ij}$ is the Kronecker delta (cf.\,\cite{petersen2008matrix} for useful matrix derivatives).

\bibliographystyle{IEEEtran}
\bibliography{IEEEabrv,bibliography}

\end{document}